\documentclass[conference]{IEEEtran}
\usepackage{amsmath,amssymb,amsfonts}
\usepackage{graphicx}
\usepackage{textcomp}
\usepackage{xcolor}
\usepackage{algorithm2e}
\usepackage{array} 
\begin{document}

\title{Optimizing Delivery Logistics: Enhancing Speed and Safety with Drone Technology}

\author{\IEEEauthorblockN{Maharshi Shastri}
\IEEEauthorblockA{Department of Computer Engineering\\Shree L.R. Tiwari College of Engineering\\Thane,India\\maharshishastri18@gmail.com}
\and
\IEEEauthorblockN{Ujjval Shrivastav}
\IEEEauthorblockA{Department of Computer Engineering\\Shree L.R. Tiwari College of Engineering\\Thane,India\\ujjvalshrivastav@gmail.com}}
\maketitle

\begin{abstract}
The increasing demand for fast and cost-effective last-mile delivery solutions has catalyzed significant advancements in drone-based logistics. This research describes the development of an AI-integrated drone delivery system, focusing on route optimization, object detection, secure package handling, and real- time tracking. The proposed system leverages YOLOv4-Tiny for object detection, the NEO-6M GPS module for navigation, and the A7670 SIM module for real-time communication. A comparative analysis of lightweight AI models and hardware components is conducted to determine the optimal configuration for real-time UAV-based delivery. Key challenges—including battery efficiency, regulatory compliance, and security considerations— are addressed through the integration of machine learning techniques, IoT devices, and encryption protocols. Preliminary studies demonstrate improvement in delivery time compared to conventional ground-based logistics, along with high-accuracy recipient authentication through facial recognition. The study also discusses ethical implications and societal acceptance of drone deliveries, ensuring compliance with FAA, EASA and DGCA regulatory standards. 
Note: This paper presents the architecture, design, and preliminary simulation results of the proposed system. Experimental results, simulation benchmarks, and deployment statistics are currently being acquired. A comprehensive analysis will be included in the extended version of this work.
\end{abstract}


\section{Introduction}
The rapid growth of e-commerce and on-demand delivery services has been amplified the demand for effacious, cost-effective, and time-sensitive logistics solutions. The traditional delivery methodologies, reliant on ground-based transportation, frequently encounter challenges such as traffic congestion, elevated operational expenditures, and ecological considerations attributed to fuel consumption and carbon emissions \cite{kumar2023urban}\cite{wahab2017comprehensive}. To mitigate these challenges, Unmanned Aerial Vehicles (UAVs), commonly designated as drones, have emerged as a disruptive technology in last-mile logistics, providing expedited delivery, diminished human intervention, and enhanced accessibility to geographically remote locations \cite{benarbia2022literature}\cite{aurambout2019last}. Recent advancements in AI-driven navigation, object detection, and IoT-based tracking have substantially improved the viability of drone deliveries. AI models such as YOLOv4-Tiny and MobileNet \cite{cantero2022benchmarking}\cite{islam2019deep} facilitate real-time object recognition, while GPS modules (e.g., NEO-6M, Ublox M8N) and SIM-based communication modules (e.g., A7670, SIM800L) \cite{spek2009sensing}enable accurate tracking and autonomous flight control. Notwithstanding these technological advancements, the integration of drones into existing logistics networks poses several challenges, including regulatory constraints, flight stability concerns, security vulnerabilities, and battery limitations. This research proposes an AI-integrated drone delivery system that optimizes route planning, security measures, and package authentication while ensuring adherence to FAA, EASA, and DGCA regulations \cite{easa2023regulatory}\cite{faa2025package}\cite{dronelaws2025india}. The system incorporates ESP32-CAM for facial recognition-based recipient verification, an A7670 SIM module for real-time tracking, and adaptive flight path algorithms for energy-efficient navigation \cite{wood2021flying}\cite{alali2024drone}. Through a comparative analysis of hardware components and AI models, this study evaluates the system's efficacy in enhancing delivery speed and security while curtailing operational costs.
This research focuses on the development of a fully autonomous drone delivery system by integrating AI-based navigation, real-time object detection, and IoT communication to enable intelligent and responsive flight control. To support effective onboard decision-making under constrained resources, the study conducts a comparative analysis of lightweight AI models—including YOLOv4-Tiny, MobileNet, and EfficientDet—to identify the optimal balance between detection accuracy and processing speed for deployment on the ESP32-CAM platform. In parallel, it explores the performance differences between two widely used GPS modules, NEO-6M and Ublox M8N, with the aim of optimizing real-time tracking, path planning, and overall navigational reliability. Recognizing the importance of legal and operational safety, the system is designed to adhere to current UAV regulations set by authorities such as the FAA, DGCA, and local aviation bodies. Furthermore, the research addresses critical operational concerns by evaluating the drone’s energy efficiency, identifying potential security vulnerabilities—especially those linked to QR-code-based verification \cite{krombholz2014qr}\cite{alzahrani2020security} and improving recipient authentication accuracy through facial recognition and color-coded identification. Collectively, these objectives aim to advance the design of practical, scalable, and secure AI-powered drone logistics systems suitable for modern urban and institutional environments.

\section{Methodology}
The proposed system presents a fully autonomous drone delivery framework designed to operate within complex urban or institutional environments, such as universities and corporate campuses, considering the airspace regulations and payload limitations\cite{narang2018india}\cite{elshafie2024comprehensive}. At its core, the drone integrates AI-based object detection , GPS-enabled navigation, real-time communication, speech-to-text recognition, and an authentication system. 
For obstacle and recipient detection, the system deploys YOLOv4-Tiny on the ESP32-CAM module due to its optimal balance of speed and accuracy under resource constraints. Navigation is facilitated through the NEO-6M GPS module, supported by the MPU6050 IMU for stabilization, enabling precise flight control and dynamic orientation adjustments. Real-time data exchange and location tracking are achieved using the A7670 SIM module, which enables cloud-based communication while maintaining cost-effective long-distance connectivity compared to traditional Wi-Fi systems.

\subsection{Steps of Delivery}
The proposed drone delivery system integrates multiple AI, IoT, and embedded technologies to enable precise, secure, and autonomous last-mile delivery, particularly suited for large campuses, corporate parks, and high-rise residential complexes. The complete delivery workflow is structured as follows:
\begin{center}
Step I.		\textbf{Recipient Order Placement} 
\end{center}
The delivery process begins with the recipient accessing in the system through a mobile application or web portal. New users are required to create an account, upon which they are assigned a unique color code displayed on their dashboard. The recipient must print and affix this color code to their designated door or collection point. 
Upon account creation, the recipient proceeds to place an order by inputting their delivery address, either automatically via the mobile device’s integrated GPS module or through speech-to-text input via the device’s microphone, utilizing models like Wave2Vec 2.0 or DeepSpeech for transaction \cite{reddy2023speech}. During this process, the recipient, is also required to upload a photograph of the authorized recipient(s) for facial verification purposes. 
Once the checkout and payment process are completed, the system generates a unique alphanumeric deliveryID, linking the recipient’s delivery address, color code, and facial image.

\begin{center}
Step II.	\textbf{Depot worker delivery initialization}
\end{center}
At the logistics depot, upon parcel arrival, a depot operator initiates the delivery sequence by inputting the assigned deliveryID into a mobile application interface. Delivery details can be entered via text or using speech-to-text capabilities \cite{reddy2023speech}. After verification, the worker places the parcel into the drone’s secure container and triggers the drone to initiate the autonomous delivery sequence.
\begin{center}
Step III. \textbf{Drone departure and flight initialization }
\end{center}
Upon activation, the drone autonomously and simultaneously: 1)Secures the container via servo-driven locking mechanism, 2)Establishes a connection with a cloud server to retrieve essential delivery parameters (color code, coordinates), 3)Converts the recipient address into latitude-longitude coordinates using the OpenStreetMap API \cite{haklay2008openstreetmap}. 
The drone then autonomously launches towards the designated location. 
\begin{center}
    Step IV.	\textbf{In-Flight operations and real-time processing}
\end{center}
During flight, the drone simultaneously executes multiple concurrent processes, necessitating a multi-core processing unit (Raspberry Pi Zero 2 W) capable of handling real-time tasks, including: 1)Continuous obstacle detection: The ESP32-CAM, operating YOLOv4-Tiny, MobileNet, or EfficientDet models \cite{ji2021real}\cite{wang2019robust}, provides real-time object detection, identifying dynamic and static obstacles; 2)Dynamic Obstacle Overpass: Upon obstacle detection, the Raspberry Pi calculates the object’s height and commands the drone to ascend an additional 5 meters to safely clear the obstacle;3)Flight stability and Navigation: The MPU6050 IMU module ensures stabilization, positioning, and precise landing control;4)Location tracking: The Neo-6M GPS module continuously feeds positional data to the system for navigation adjustments \cite{he2025research};5)Cloud-based communication: The A7670 SIM module facilitates cost-effective, long-range communication by uploading the drone’s location and telemetry data in real-time.
All these operations are orchestrated concurrently via multi-threading on the Raspberry Pi Zero 2 W until the drone reaches a 6-meter radius of the recipient’s location. 
\begin{figure}[htbp]
    \centering
    \includegraphics[width=8.9cm]{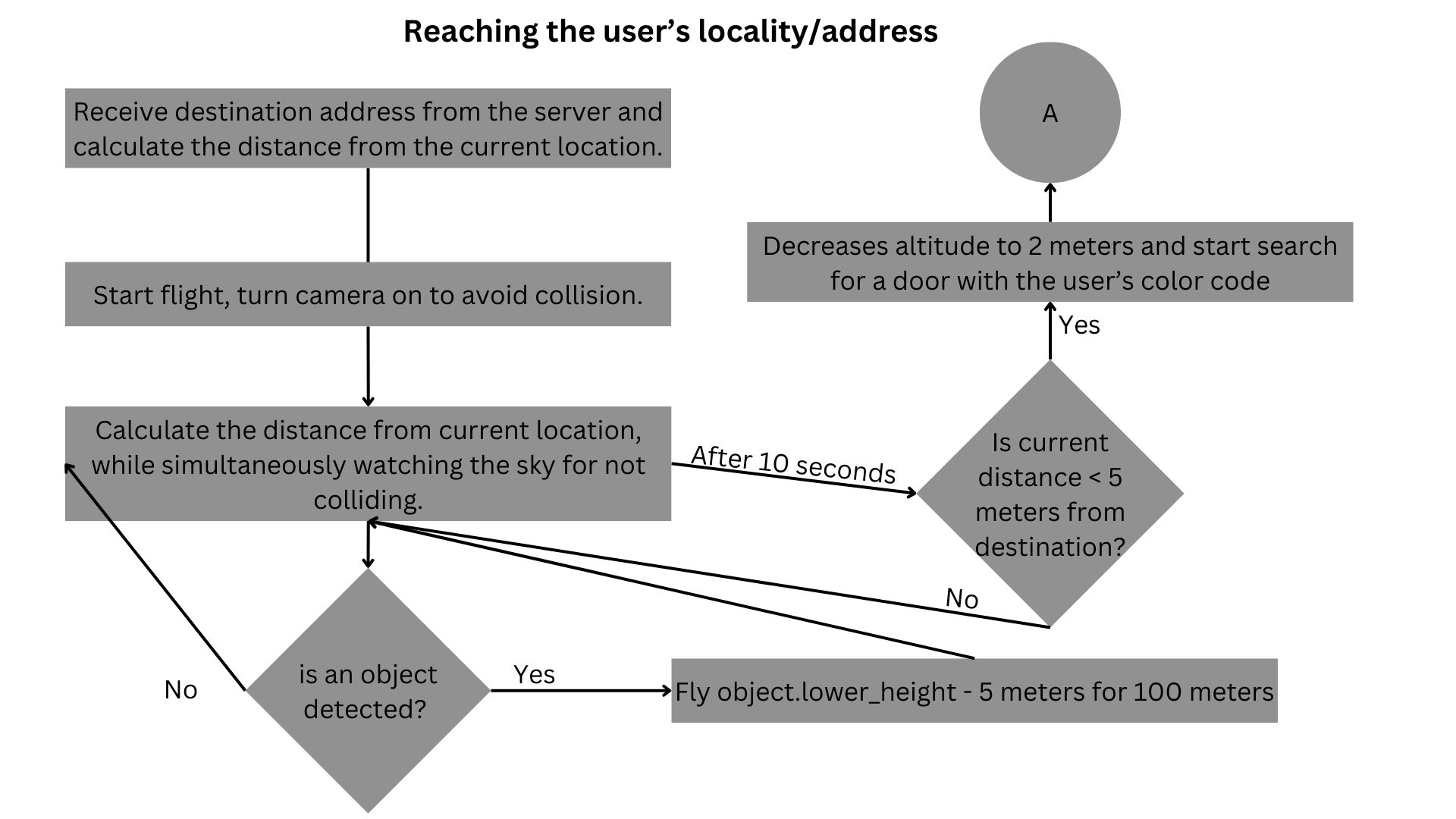}
    \caption{\textbf{Reaching the user’s locality via the process explain in Step IV, in a flowchart}}
\end{figure}
\begin{center}
    Step V.	\textbf{Target Localization via Color code detection}
\end{center}
Upon reaching the vicinity of the recipient’s address, the drone descends to an altitude of approximately 2 meters and initiates a visual scanning process to identify the door displaying the assigned color code. Using YOLOv4-Tiny deployed on the ESP32-CAM, the drone sequentially scans all the doors of the building. Algorithm(presented in pseudo code) of identifying the recipient’s door:
\begin{algorithm}
\caption{Function \texttt{recipient\_door\_recognition(color\_code)}}\label{alg:cap}
\SetKwInput{KwBegin}{Begin} 
\SetKwInput{KwEnd}{End}     

\KwBegin{} 
    Enter the building
    \While{not all doors of the building are scanned}{
        Select the next unscanned door
        Scan the door for user's color code
        \If{user's color code matches}{
            Push notification to user: "Accept Delivery"
            Request user's image file to server
            \textbf{break} 
        }
        Push notification to user: "Missed delivery"
        Re-direct to depot
    }
\KwEnd{} 
\end{algorithm}
\begin{figure}[htbp]
    \centering
    \includegraphics[width=8.9cm]{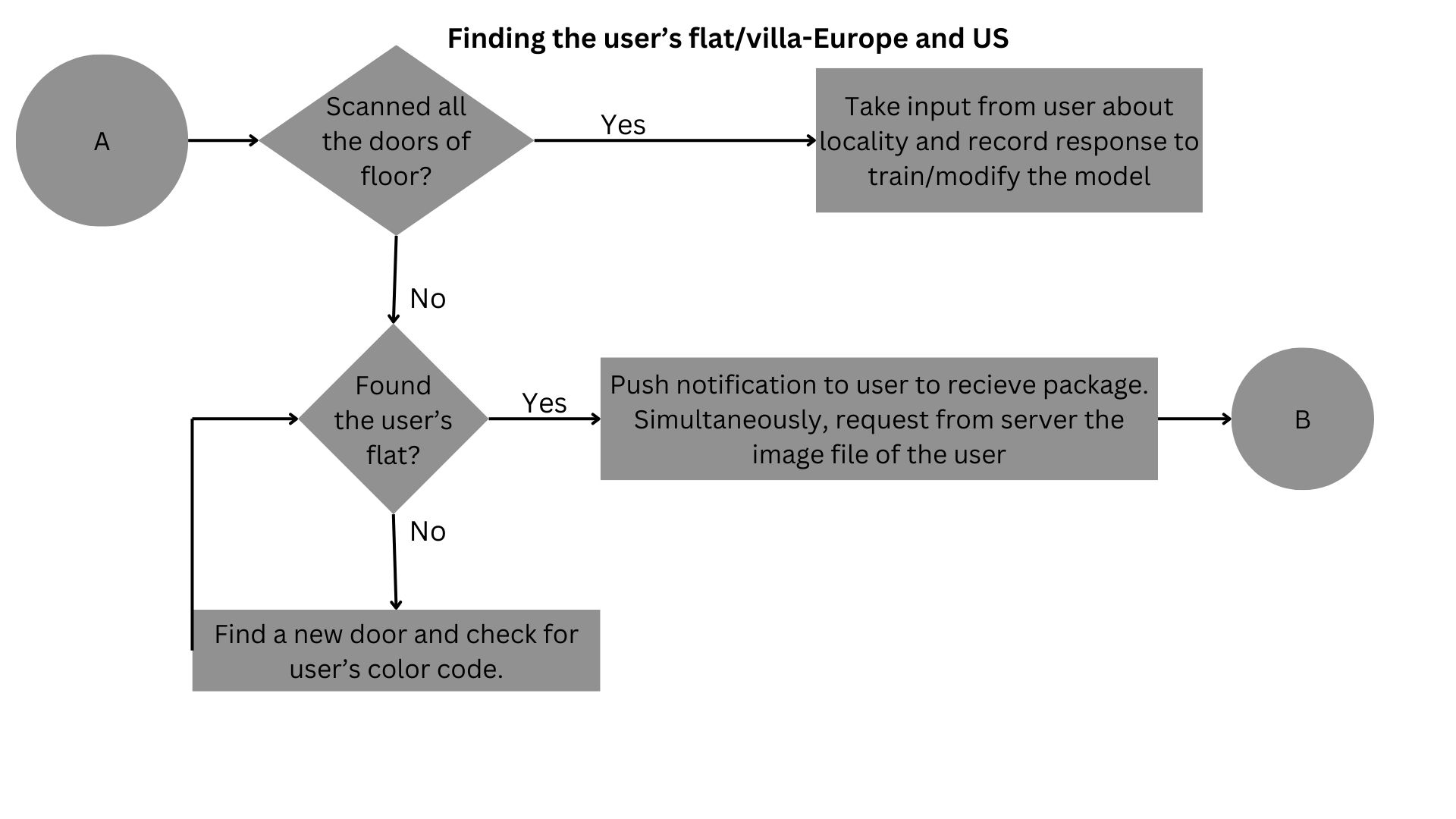}
    \caption{\textbf{Reaching the designated flat of the user via the process explained in Step V, in a  flowchart}}
\end{figure}
\\Upon reaching the recipient's door, the system transitions to the delivery verification phase.
\begin{center}
    Step VI. \textbf{Secure delivery and Recipient Authentication}
\end{center}
Following successful localization, the drone will run the following algorithm (presented here in pseudo code)
\begin{algorithm}[ht]
\caption{Function ConfirmDelivery}\label{alg:cap}
\SetKwInput{KwBegin}{Begin} 
\SetKwInput{KwEnd}{End}     

\KwBegin{} 
    Receive recipient’s reference image from the server;
    $image\_file \gets save("image.jpg");$
    $start\_time \gets current\_time;$
    \While{$delivery\_status is not determined$}{
        Livestream and wait for a face to be detected via ESP32-CAM;
        Perform face recognition (match live image with received image);

    \If{current\_time - start\_time $\geq$ 600}{
        $message \gets "Not delivered"$;
        \textbf{break}};
        \ElseIf{$face matches with "image.jpg" \ge 0.8$}{
            $servo\_message \gets "open";$
            sleep(30);
            $message \gets "Delivered";$

        \textbf{break}};
        \Else{
        continue monitoring for face;
            }}
    Depart for depot;
    Return message;
\KwEnd{}
\end{algorithm}
\hspace{1 cm}During this window, the ESP32-CAM actively scans the approaching individual. Upon achieving facial recognition \cite{badashah2022automatic} \cite{malovic2024cost} confidence level exceeding 80\%, the servo motor unlocks the container, allowing the authorized recipient to retrieve the parcel.
\hspace{1 cm}If the recipient fails to authenticate within the allocated time, the drone automatically returns to depot, ensuring package security.
\hspace{1 cm}Upon successful delivery or timeout expiration, the drone concludes the mission and prepares for the next deployment.
\section{System design(hardware and software)}
\begin{figure}[htbp]
    \centering
    \includegraphics[width=8.9cm]{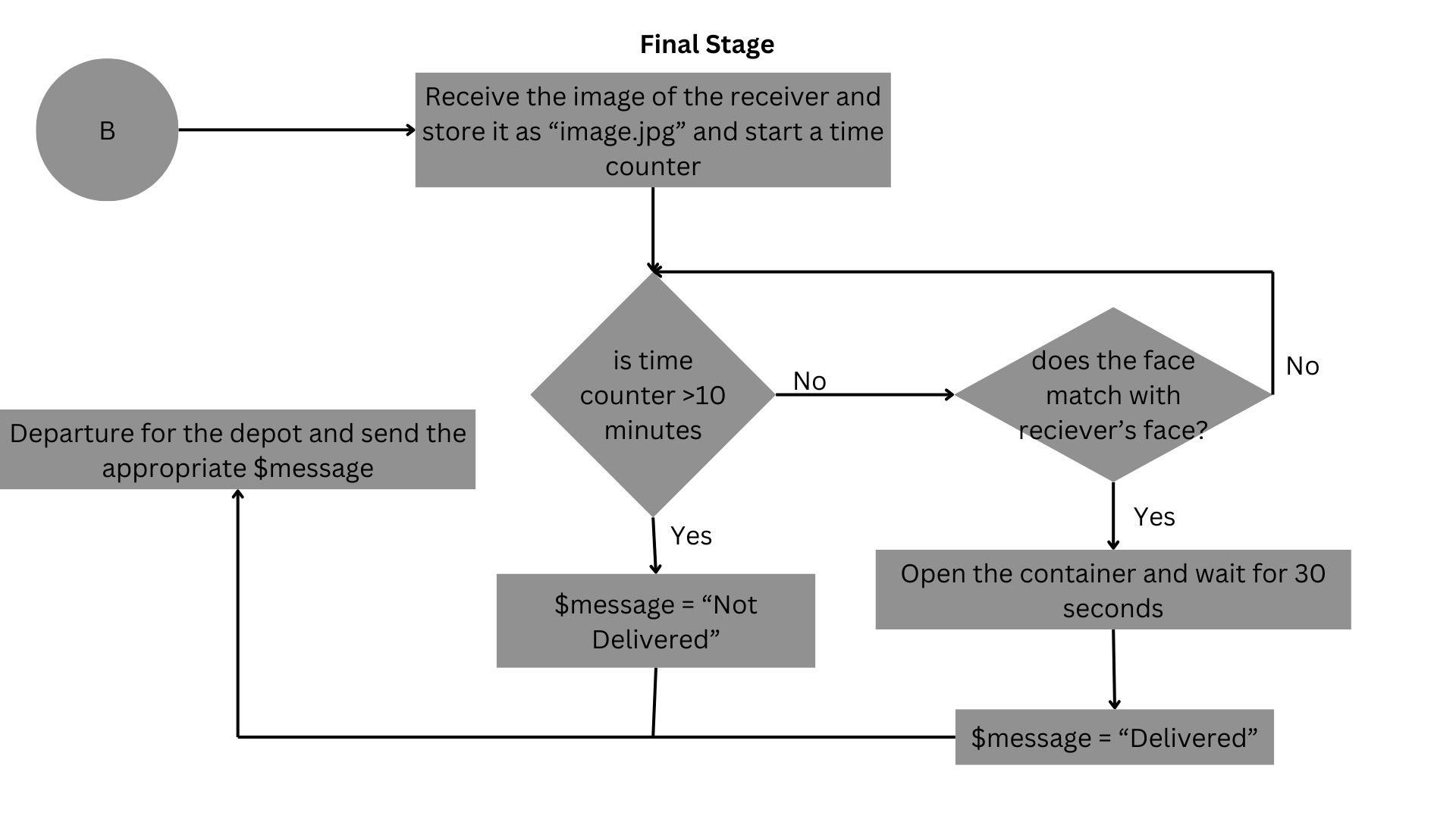}
    \caption{\textbf{Performing final verification via the process explained in Step VI, in a  flowchart}}
\end{figure}
\begin{table}
    \centering
    \begin{tabular}{|c|>{\centering\arraybackslash}p{0.35\linewidth}|c|}\hline
         Component name&  Function& Cost(INR)\\\hline
         ESP32-CAM&  Camera module used for object detection and facial recognition& 891\\\hline
         Raspberry Pi Zero 2W&  Serves as the main central processing unit of the system& 1647\\\hline
         Esp32-Wrover&  GPS and motor communication& 303\\\hline
         Neo-6M&  Provides real-time latitude, longitude and altitude& 538\\\hline
         BMP280&  Measures barometric pressure and temperature for altitude estimation& 33\\\hline
         A7670&  Enables mobile network-based communication& 1499\\\hline
         Power Distribution Board&  Distributes power safely at regulated voltages& 137\\\hline
         3S LiPo battery&  Provides power supply to all components& 1999\\\hline
         MPU6050&  Assist in flight stabilization and navigation& 199\\ \hline
         Servo Motor&  Controls opening and closing of the parcel chamber& 76\\ \hline
         Electronic Speed Controller&  Regulates brushless motor speed& 299\\ \hline
         Brushless DC motor& Propulsion motor for drone flight& 400
    \\ \hline\end{tabular}
    \caption{The table outlines the essential components and their functions, highlighting how each contributes to the system's autonomous operation.}
    \label{Table 1:Table of components}
\end{table}
\textbf{Architecture diagram of the overall hardware and proposed system}

The architecture diagram illustrates the end-to-end workflow of the proposed autonomous drone delivery system, combining both hardware-level integration and high-level operational logic. At its core, the diagram maps the interaction between key hardware modules—including the Raspberry Pi Zero 2 W, ESP32-CAM, GPS, IMU, and communication modules—with various subsystems responsible for navigation, object detection, flight control, and parcel handling. The system architecture further visualizes the cloud communication loop, beginning with order initiation by the recipient through a mobile/web interface, followed by the transmission of delivery metadata—such as GPS coordinates, facial image, and color code—to the drone via cloud synchronization. The diagram also captures the decision-making logic, including obstacle detection and avoidance, altitude adjustments, color code verification, and facial recognition-based delivery. This architectural overview is critical in demonstrating how hardware components and software processes are seamlessly orchestrated to achieve real-time autonomy, secure delivery, and system scalability.\\
\begin{figure}[htbp]
    \centering
    \includegraphics[width=8.9cm]{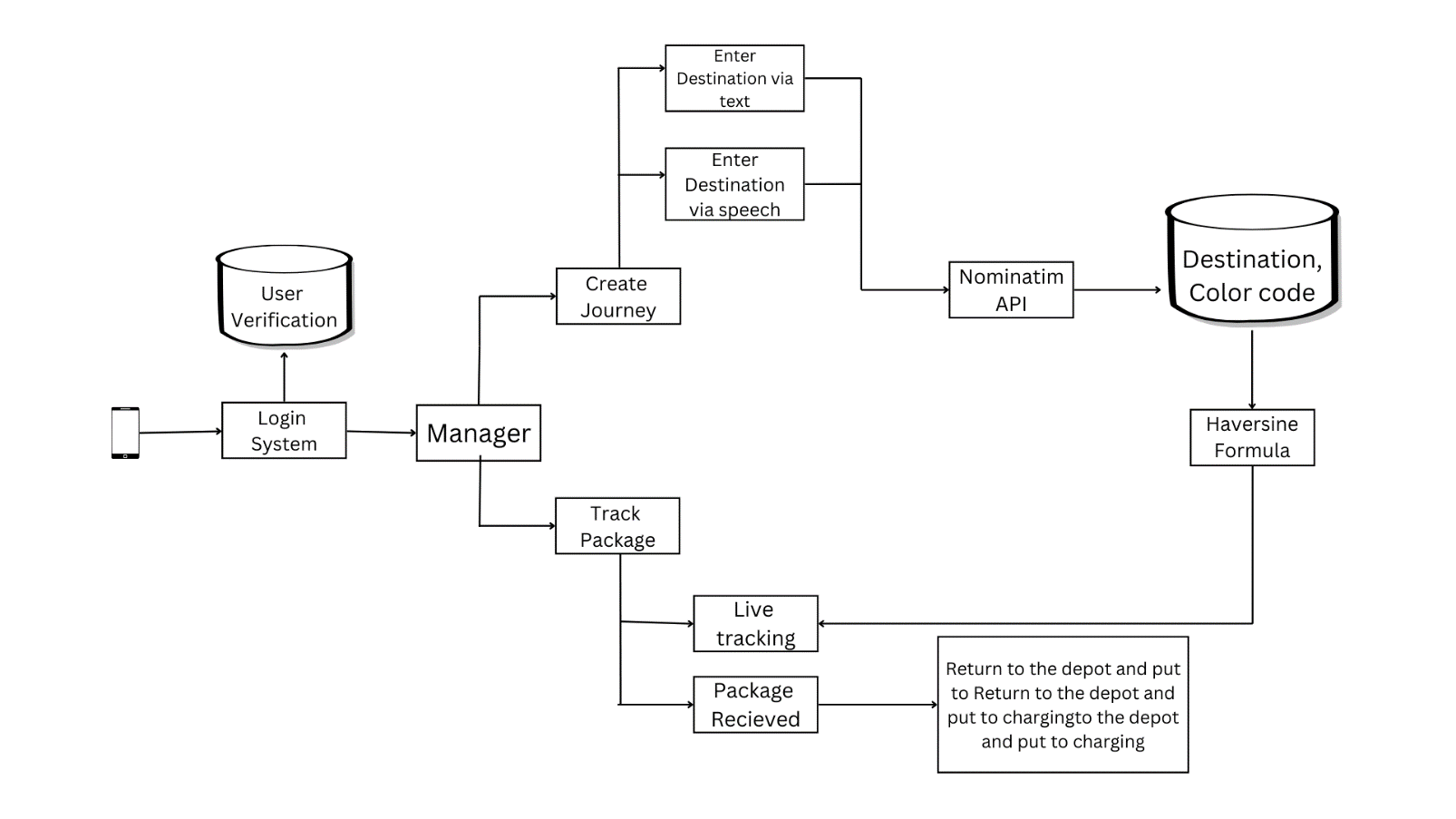}
    \caption{\textbf{: System architecture depicting the end-to-end workflow of the autonomous drone delivery process.}}
\end{figure}
 \textbf{Figure 4} represents the complete operational flow of the drone delivery system, beginning with the user login and verification process, which authenticates the recipient through a secure interface. Once verified, the user proceeds to create a delivery journey by entering the destination address either via text input or speech, the latter being processed through a speech-to-text system. This destination data is converted to geographic coordinates using the Nominatim API, while simultaneously retrieving the associated color code for door identification.\\
The Manager module acts as the central coordinator, overseeing the creation of the delivery route, tracking the package, and enabling live tracking throughout the drone’s journey. The Haversine formula is used to calculate the real-time distance between the drone’s current location and the delivery address to optimize flight decisions. Upon successful delivery and recipient verification, the system logs the status as “Package Received.” The drone then automatically enters a return sequence to the depot, where it is either prepared for the next delivery cycle or directed to a charging station, ensuring continued autonomous operation without human intervention.
\begin{figure}[htbp]
    \centering
    \includegraphics[width=8.9cm]{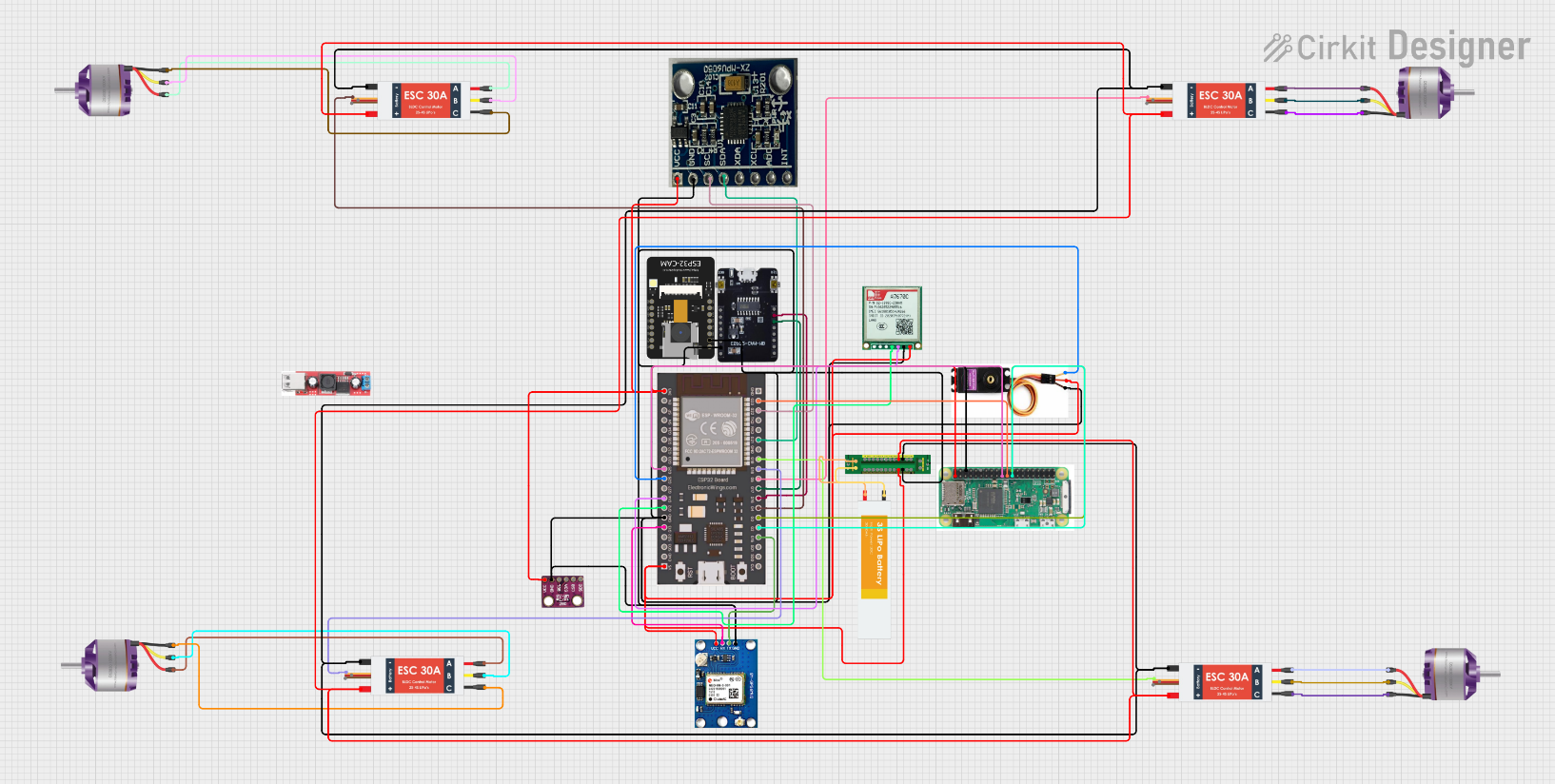}
    \caption{\textbf{Hardware architecture of the drone with connections with each component}}
\end{figure}
\textbf{Figure 5} illustrates the integration of multiple components; each connected through specific interfaces to ensure seamless communication and operation. The Raspberry Pi Zero 2W serves as the central processing unit, interfacing with various sensors and modules via GPIO pins. The ESP32-CAM handles real-time transmission of data to the Raspberry Pi via serial communication. Flight stability is maintained through the MPU6050 IMU, which provides accelerometer and gyroscope data, communicated to the Pi Zero via I2C. The Neo-6M GPS module, connected to the ESP32, continuously provides the drone’s geographic coordinates, aiding in navigation. BMP280 sensors measure atmospheric pressure for altitude estimation, while cloud connectivity is enabled via the A7670C module, which communicates with the Pi Zero using UART. The drone’s propulsion system is powered by Brushless DC motors, controlled through Electronic Speed Controllers (ESCs) connected to the ESP32, with motor speed adjustments managed via PWM signals. A servo motor is integrated to open and close the drone’s parcel chamber, controlled by the ESP32 through PWM input. Communication between the ESP32 and Pi Zero is facilitated through SPI, enabling efficient data exchange for autonomous operation and real-time processing. The entire system is powered by a 3S LiPo battery, regulated by a power distribution board to ensure stable voltage levels across all components. This architecture provides a robust and efficient setup for autonomous drone delivery, enabling precise control, navigation, and parcel handling.   
\section{Results and Discussions}
Results and pending. Experimental simulations in progress. Will be updated in v2.
\bibliographystyle{IEEEtran}
\bibliography{references.bib}  
\end{document}